\pdfoutput=1

\documentclass[11pt]{article}

\usepackage[]{EMNLP2022}

\usepackage{times}
\usepackage{balance}
\usepackage{latexsym}
\usepackage{graphicx}
\usepackage{microtype}
\usepackage{booktabs}
\usepackage{multirow}
\usepackage{pifont}
\usepackage{bm}
\usepackage{url}
\usepackage{algorithm}
\usepackage{algorithmicx}
\usepackage{algpseudocode}
\usepackage{amsmath}
\usepackage{makecell}
\usepackage{bbding}
\usepackage{fancyhdr}
\usepackage{longtable}
\usepackage{amssymb}
\usepackage{stmaryrd,pict2e,picture}

\usepackage[T1]{fontenc}

\usepackage[utf8]{inputenc}

\usepackage{microtype}

\usepackage{inconsolata}

\newtheorem{lemma}{Lemma}

\title{LightEA: A Scalable, Robust, and Interpretable Entity Alignment Framework via Three-view Label Propagation}

\author{Xin Mao$^{1,2}$, Wenting Wang$^3$, Yuanbin Wu$^{1,2}$, Man Lan$^{1,2}$\\
  $^1$School of Computer Science and Technology, East China Normal University \\
  $^2$Shanghai Institute of AI for Education, East China Normal University \\
  $^3$TikTok Group, Singapore\\
  \texttt{xmao@stu.ecnu.edu.cn,\{ybwu,mlan\}@cs.ecnu.edu.cn} \\ \texttt{wenting.wang@bytedance.com}}

\begin{document}
\maketitle
\begin{abstract}
Entity Alignment (EA) aims to find equivalent entity pairs between KGs, which is the core step of bridging and integrating multi-source KGs.
In this paper, we argue that existing GNN-based EA methods inherit the inborn defects from their
neural network lineage: weak scalability and poor interpretability.
Inspired by recent studies, we reinvent the \emph{Label Propagation} algorithm to effectively run on KGs and propose a non-neural EA framework --- LightEA, consisting of three efficient components: (i) \emph{Random Orthogonal Label Generation}, (ii) \emph{Three-view Label Propagation}, and (iii) \emph{Sparse Sinkhorn Iteration}.
According to the extensive experiments on public datasets, LightEA has impressive scalability, robustness, and interpretability.
With a mere tenth of time consumption, LightEA achieves comparable results to state-of-the-art methods across all datasets and even surpasses them on many.
\end{abstract}

\section{Introduction}
\label{sec:intro}

Knowledge Graph (KG) describes the real-world entities and their internal relations by triples $(head, rel, tail)$, expressing the information on the internet in a form closer to human cognition.
To this day, KGs have facilitated a mount of downstream internet applications (e.g., search engines \cite{DBLP:conf/icde/YangAJTW19} and dialogue systems \cite{DBLP:conf/emnlp/YangZE20}) and become one of the core driving forces in the development of artificial intelligence.
In practice, KGs are usually constructed by various departments with multi-source data.
Therefore, they typically contain complementary knowledge while having overlapping parts.
Integrating these independent KGs could significantly improve the coverage rate, which is especially beneficial to low-resource language users.

\begin{figure}
    \centering
    \includegraphics[width = 0.95\linewidth]{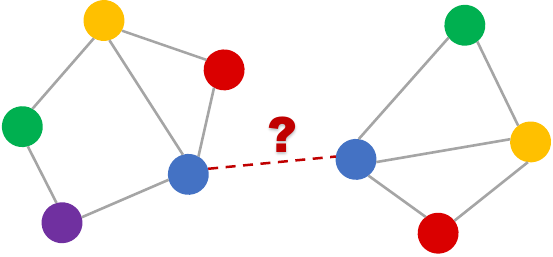}
    \caption{A toy example of entity alignment.}
    \label{fig:intro}
\end{figure}

Entity Alignment (EA) aims to find equivalent entity pairs between KGs (as shown in Figure \ref{fig:intro}), which is the core step of bridging and integrating multi-source KGs.
Therefore, EA attracts enormous attention and progresses rapidly.
Most existing methods regard EA as a graph representation learning task and share the same two-stage architecture: (i) encoding the KGs into low-dimensional spaces via graph encoders (e.g., TransE \cite{DBLP:conf/nips/BordesUGWY13} and GCN \cite{DBLP:journals/corr/KipfW16}) and (ii) mapping the embeddings of equivalent entity pairs into a unified vector space through contrastive losses \cite{DBLP:conf/cvpr/HadsellCL06}.

Recently, \emph{Graph Neural Network} (GNN) has achieved impressive success in many sorts of graph applications.
Following \citet{DBLP:conf/emnlp/WangLLZ18}, who first introduced \emph{Graph Convolutional Network} (GCN) into EA, numerous new fancy mechanisms are proposed and stacked over vanilla GCN for better performance, such as \emph{Graph Matching} \cite{DBLP:conf/iclr/FeyL0MK20}, \emph{Relational Attention} \cite{DBLP:conf/wsdm/MaoWXLW20}, and \emph{Hyperbolic embedding} \cite{DBLP:conf/emnlp/SunCHWDZ20}.
According to an EA paper list \footnote{\url{github.com/THU-KEG/Entity_Alignment_Papers}} on Github, over $90\%$ of EA methods adopted GNNs as their graph encoders in recent three years.

However, there is no such thing as a free lunch.
These increasingly complex GNN-based methods inherit the following inborn defects from their GNN lineage:
(i) \textbf{Weak scalability}.
Since the scales of real-world graphs are usually massive (e.g., YAGO$3$ \cite{DBLP:conf/www/SuchanekKW07} contains $17$ million entities), the scalability of graph algorithms is crucial.
However, as summarized by \citet{9174835}, most advanced EA methods require several hours \cite{DBLP:conf/ijcai/SunHZQ18, DBLP:conf/acl/CaoLLLLC19} or even days \cite{DBLP:conf/acl/XuWYFSWY19} on the DWY$100$K dataset, which only contains $200,000$ nodes.
Although an efficient loss function \cite{DBLP:conf/www/MaoWWL21} or a graph sampler \cite{DBLP:journals/corr/abs-2205-10312} could effectively alleviate this problem, all existing EA methods are still overstretched when facing the real-world KGs.
(ii) \textbf{Poor interpretability}.
Interpreting neural networks is a recognized challenge, and the complex graph structure makes it more difficult.
A few studies try to explain the behaviors of EA methods by showing wrong cases \cite{DBLP:conf/emnlp/YangZSLLS19} or visualizing attention weights \cite{DBLP:conf/acl/WuLFWZ20}.
And most EA studies \cite{DBLP:conf/ijcai/SunHZQ18, DBLP:conf/acl/XuWYFSWY19, DBLP:conf/cikm/MaoWXWL20} do not attempt to give any interpretation, only focusing on improving the performances on evaluation metrics.

An ancient Chinese saying goes, \emph{"drawing new inspiration while reviewing the old."}
A recent study \cite{DBLP:conf/iclr/HuangHSLB21} reinvents the classical graph algorithm --- \emph{Label Propagation} (LP) \cite{Zhu2002LearningFL}, combining it with shallow neural networks.
Surprisingly, this simple method out-performs the current best-known GNNs with more than two orders of magnitude fewer parameters and more than two orders of magnitude less training time.
Inspired by their excellent work, this paper proposes the \emph{Three-view Label Propagation} mechanism that enables the LP algorithm, designed for homogeneous graphs, to effectively run on KGs (a kind of typical heterogeneous graph).
Besides, we further propose two approximation strategies to reduce the computational complexity and enhance the scalability:
\emph{Random Orthogonal Label Generation} and \emph{Sparse Sinkhorn Iteration}.
The above three components constitute the proposed non-neural EA framework --- LightEA.
According to the extensive experiments on four groups of public datasets, LightEA has impressive scalability, robustness, and interpretability:

(1) \textbf{Scalability}:
Unlike GNN-based EA methods that require multiple rounds of forward and backward propagation, LightEA only requires one round of label propagation without any trainable parameters.
After abandoning neural networks, LightEA achieves extremely high parallel computing efficiency.
With a PC that has one RTX$3090$ GPU, LightEA only takes $7$ seconds to obtain the alignment results on DBP$15$K and less than $35$ seconds on DWY$100$K, which is only one-tenth of the state-of-the-art EA method.
LightEA could also easily handle DBP$1$M which contains more than one million entities and nearly ten million triples, while most EA methods even cannot run on it.
Besides running speed, the flexible framework enables LightEA could easily incorporate iterative strategies and literal features (e.g., entity names) to improve performance.

(2) \textbf{Robustness}:
In this paper, we design a thorough robustness examination that evaluates LightEA on four groups of public datasets containing cross-lingual, mono-lingual, sparse, dense, and large-scale subsets.
With a mere tenth of time consumption, LightEA achieves comparable results to state-of-the-art methods across all datasets and even surpasses them on many.
Besides, since LightEA does not have trainable parameters, the performance fluctuation of multiple runs is limited.

(3) \textbf{Interpretability}:
Researchers generally consider linear models (e.g., Linear Regression) have decent interpretability because their outputs are the linear summation of the input features.
Consistent with the LP algorithm, the computational process of LightEA is also entirely linear.
After removing \emph{Random Orthogonal Label}, each dimension of the label vectors will have a clear and realistic meaning.
We could trace the propagation process at each step to clearly explain how the entities are aligned.

In addition to the above contributions, we further design extensive auxiliary experiments to investigate the behaviors of LightEA in various situations.
The source code and datasets are now available in Github (\url{github.com/MaoXinn/LightEA}).

\section{Task Definition}
A KG could be defined as $\mathcal{G} = (\mathcal{E},\mathcal{R},\mathcal{T})$, where $\mathcal{E}$ is the entity set, $\mathcal{R}$ is the relation set, and $\mathcal{T}\subset \mathcal{E}\times \mathcal{R}\times \mathcal{E}$ represents the set of triples.
Given the source graph $\mathcal{G}_s$, the target graph $\mathcal{G}_t$, and the set of pre-aligned entity pairs $\mathcal{P}$, EA aims to find new equivalent entity pairs based on $\mathcal{G}_s$, $\mathcal{G}_t$, and $\mathcal{P}$.

\section{Related Work}
\subsection{Entity Alignment}
Most existing methods regard EA as a graph representation learning task and share the same two-stage architecture: (i) encoding the KGs into low-dimensional spaces via graph encoders and then (ii) mapping the embeddings of equivalent entity pairs into a unified vector space by contrastive losses.

Graph encoder is the most prominent and important part of existing EA methods.
Early methods usually used TransE \cite{DBLP:conf/nips/BordesUGWY13} and its variants as the graph encoder.
However, due to TransE only focusing on optimizing independent triples $\bm h + \bm r \approx \bm t$, it lacks the ability to model the global structure of KGs.
With impressive capability in modeling graph data, GNNs quickly become the mainstream algorithm for almost all graph applications, including entity alignment.
Since \citet{DBLP:conf/emnlp/WangLLZ18} first introduced GCN into EA, numerous GNN-based EA methods have been springing up.
For example, GM-Align \cite{DBLP:conf/acl/XuWYFSWY19} introduces \emph{Graph Matching Networks} to capture the entity interactions across KGs.
RREA \cite{DBLP:conf/cikm/MaoWXWL20} proposes the \emph{Relational Reflection} operation to generate relation-specific entity embeddings.
HyperEA \cite{DBLP:conf/emnlp/SunCHWDZ20} adopts hyperbolic embeddings to reduce the dimension of entities.

Besides the modifications on encoders, some EA methods adopt iterative strategies to generate semi-supervised data due to the lack of labeled entities.
Some EA methods propose that introducing literal information (e.g., entity names) could provide a multi-aspect view for alignment models.
However, it should be noted that not all KGs contain literal information, especially in practical applications.
Table \ref{tabel:rw} categorizes some popular EA methods based on their encoders and whether using iterative strategies or literal information.

\begin{table}
\resizebox{1\linewidth}{!}{
\renewcommand\arraystretch{0.9}
\begin{tabular}{cccc}
  \toprule
  \textbf{Method}&\textbf{Encoder}&\textbf{Literal}&\textbf{Iterative}\\
  \midrule
  MtransE\cite{DBLP:conf/ijcai/ChenTYZ17} &Trans&&\\
  JAPE \cite{DBLP:conf/semweb/SunHL17}&Trans&&\\
  BootEA \cite{DBLP:conf/ijcai/SunHZQ18}&Trans&&\CheckmarkBold\\
  KDCoE \cite{DBLP:conf/ijcai/ChenTCSZ18}&Trans&\CheckmarkBold&\CheckmarkBold\\
  RSN \cite{DBLP:conf/icml/GuoSH19} &RNN&&\\
  TransEdge\cite{DBLP:journals/corr/abs-2004-13579}&Trans&&\CheckmarkBold\\
  \midrule
  GCN-Align \cite{DBLP:conf/emnlp/WangLLZ18}&GNN&&\\
  MuGNN \cite{DBLP:conf/acl/CaoLLLLC19}&GNN&&\\
  RDGCN \cite{DBLP:conf/ijcai/WuLF0Y019}&GNN&\CheckmarkBold&\\
  GM-Align \cite{DBLP:conf/acl/XuWYFSWY19}&GNN&\CheckmarkBold&\\
  HyperKA \cite{DBLP:conf/emnlp/SunCHWDZ20}&GNN&&\\
  MRAEA \cite{DBLP:conf/wsdm/MaoWXLW20}&GNN&&\CheckmarkBold\\
  RREA \cite{DBLP:conf/cikm/MaoWXWL20}&GNN&&\CheckmarkBold\\
  Dual-AMN \cite{DBLP:conf/www/MaoWWL21}&GNN&&\CheckmarkBold\\
  EASY \cite{DBLP:conf/sigir/GeLCZG21}&GNN&\CheckmarkBold&\CheckmarkBold\\
  ClusterEA \cite{DBLP:journals/corr/abs-2205-10312}&GNN&&\CheckmarkBold\\
  \bottomrule
\end{tabular}
}
\caption{Categorization of some popular EA methods.}\label{tabel:rw}
\end{table}

\subsection{Label Propagation and GCN}
\emph{Label Propagation} (LP) \cite{Zhu2002LearningFL} is a classical graph algorithm for node classification and community detection.
It assumes that two connected nodes are likely to have the same label, and thus it propagates labels along the edges.
Let $\bm A \in \mathbb{R}^{|\mathcal{E}|\times |\mathcal{E}|}$ be the graph adjacency matrix, $\bm D \in \mathbb{R}^{|\mathcal{E}|\times |\mathcal{E}|}$ be the diagonal degree matrix of $\bm A$, $\bm L^{(k)} \in \mathbb{R}^{|\mathcal{E}|\times c}$ be the label matrix after $k$ rounds of label propagation, where $c$ is the number of classes.
Each column of $\bm L^{(0)}$ is a one-hot vector, initialized by the input labels of the known nodes, while the label vectors of unknown nodes remain zeros.
The propagation process of the random LP algorithm could be formulated as follow:
\begin{equation}
    \bm L^{(k+1)}=\bm D^{-1}\bm A\bm L^{(k)}\\
\end{equation}

\emph{Graph Convolutional Network} (GCN) \cite{DBLP:journals/corr/KipfW16} is a multi-layer neural network that propagates and transforms node features across the graph.
Let $\bm H^{(k)} \in \mathbb{R}^{|\mathcal{E}|\times d_{in}}$ be the input features of layer $k$, and $\bm W^{(k)} \in\mathbb{R}^{d_{in}\times d_{out}}$ be the transformation matrix.
The layer-wise propagation process of GCN could be summarized as follow:
\begin{equation}
    \bm H^{(k+1)}=\sigma(\bm D^{-1/2}\bm A\bm D^{-1/2}\bm H^{(k)}\bm W^{(k)})\\
\end{equation}

Equations (1) and (2) indicate that the common intuition behind both LP and GCN is smoothing the labels or features.
\citet{DBLP:journals/corr/abs-2002-06755} notice this kind of inner correlation and unify them into one framework.
\citet{DBLP:conf/iclr/HuangHSLB21} combine shallow neural networks with the LP algorithm, achieving comparable performances to state-of-the-art GNNs, but with a small fraction of the parameters and time consumption.
These excellent works inspire us that the LP algorithm, neglected by the EA community, deserves further investigation.

\begin{figure*}
    \centering
    \includegraphics[width=\textwidth]{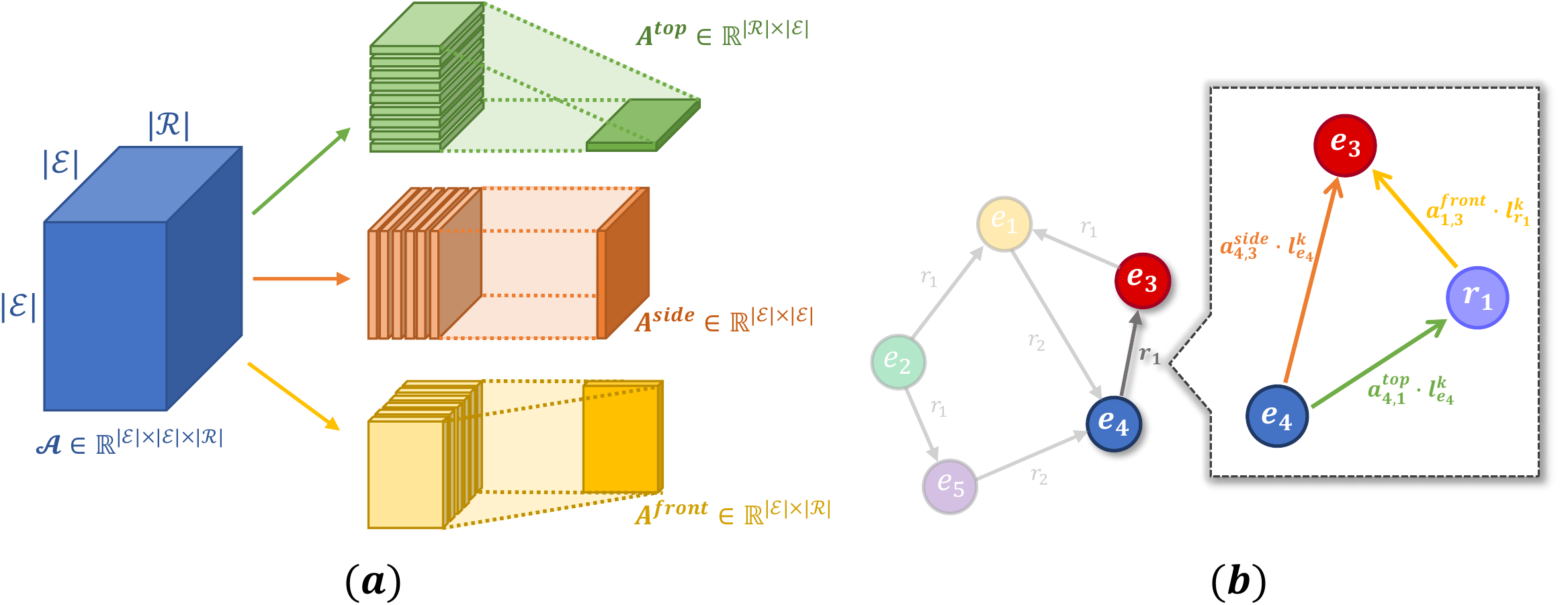}
    \caption{Illustrations of \emph{Three-view Label Propagation}.}
    \label{fig:MRW}
\end{figure*}

\section{The Proposed Method}
LightEA is a non-neural EA framework consisting of three components: (i) \emph{Random Orthogonal Label Generation}, (ii) \emph{Three-view Label Propagation}, and (iii) \emph{Sparse Sinkhorn Iteration}.
We will describe each component of LightEA in this section.

\subsection{Random Orthogonal Label Generation}
Different from the node classification and community detection tasks, the entities in EA do not have explicit class labels.
LightEA borrows a common idea from face recognition \cite{DBLP:journals/spl/WangCLL18,DBLP:conf/cvpr/DengGXZ19} that regards each pair of pre-aligned entities as a independent class.
Assume that $(e_i,e_j)\in \mathcal{P}$ is the $x$-$th$ pre-aligned entity pair, the input label matrix of entities $\bm L^{(0)}_e = \left[\bm l^{(0)}_{e_1},\bm l^{(0)}_{e_2},...,,\bm l^{(0)}_{e_{|\mathcal{E}|}}\right]$ is initialized as follows:
\begin{equation}
    \bm l^{(0)}_{e_i}=\bm l^{(0)}_{e_j}=onehot(x)\;\;\;\forall(e_i,e_j)\in \mathcal{P}
\end{equation}
where $onehot(x) \in \mathbb{R}^{|\mathcal{P}|}  $ represents the one-hot vector that only the $x$-$th$ element equals one.
The input label vectors of remaining unaligned entities are initialized to all-zero.
Besides, since existing EA datasets do not provide pre-aligned relation pairs, the input label matrix of relations $\bm L^{(0)}_r \in \mathbb{R}^{|\mathcal{R}|\times|\mathcal{P}|}$ is also initialized to all-zero.

However, this initialization strategy will cause the input label matrices of entities and relations overly large and extremely sparse.
To ensure that LightEA runs well on large-scale datasets, we have to seek a solution that could reduce the dimension of input label matrices while the loss of orthogonality is slight.
Fortunately, independent random vectors on the high-dimensional hyper-sphere could satisfy our requirement.
\begin{lemma}
If $\bm x$ and $\bm y$ are independent random unit vectors on the $d$-dimensional hyper-sphere, $\left\langle\cdot\right\rangle$ represents the inner-product operation, then we have:
\begin{equation}
    P\left(\langle\bm x,\bm y\rangle > \epsilon\right) \leq (1-\epsilon^2)^{(d+1)/2}
\end{equation}
\label{lemma:1}
\end{lemma}

Lemma \ref{lemma:1} \cite{Ball1997AnEI} states that any two independent random vectors on the high-dimensional hyper-sphere are approximately orthogonal.
For example, when $d>2048$, the probability upper bound of $\left\langle\bm x,\bm y\right\rangle > 0.1$ is less than $3.37\times 10^{-5}$.
Therefore, LightEA independently samples random vectors on the $d$-dimensional hyper-sphere to approximate the one-hot label vectors for better space-time complexity:
\begin{equation}
    \bm l^{(0)}_{e_i}=\bm l^{(0)}_{e_j}=random(d)\;\;\;\forall(e_i,e_j)\in \mathcal{P}
\end{equation}
With this approximation strategy, the dimensions of $\bm L^{(0)}_e $ and $\bm L^{(0)}_r$ are reduced to $\mathbb{R}^{|\mathcal{E}|\times d}$ and $\mathbb{R}^{|\mathcal{R}|\times d}$.

One of the significant differences between LP and GCN is that one propagates labels, and the other propagates features.
However, the above discussion indicates that randomly initialized features could be regarded as an approximation to one-hot labels.
From this perspective, the propagation processes of GCN and LP are equivalent.

\subsection{Three-view Label Propagation}
Both LP algorithm and vanilla GCN were originally designed for homogeneous graphs.
However, KG is a kind of typical heterogeneous graph that requires a third-order tensor $\boldsymbol{\mathcal{A}}\in \mathbb{R}^{|\mathcal{E}|\times|\mathcal{E}|\times|\mathcal{R}|}$ to fully describe its adjacency relations.
Some early EA methods \cite{DBLP:conf/emnlp/WangLLZ18,DBLP:conf/acl/XuWYFSWY19} crudely treat all relations as equivalent, resulting in information losses.
Follow-up studies usually address this problem by adopting the relational attention mechanism \cite{DBLP:conf/ijcai/WuLF0Y019,DBLP:conf/wsdm/MaoWXLW20} to learn attention parameters and assign different weights for different relations and triples.

Besides the above methods, an intuitive solution for generalizing LP on KGs is to use the tensor-matrix product to replace the matrix product:
\begin{align}
  \bm L^{(k+1)}_e &= \boldsymbol{\mathcal{A}} \times_2 \bm L^{(k)}_e \times_3 \bm L^{(k)}_r\\
  \bm L^{(k+1)}_r &= \boldsymbol{\mathcal{A}} \times_1 \bm L^{(k)}_e \times_2 \bm L^{(k)}_e
\end{align}
where $\times_i$ represents the $i$-mode tensor-matrix product (i.e., along the $i$-$th$ axis).
Unfortunately, this solution has two fatal defects:
(i) The tensor-matrix product leads to a squared increase in the dimension after each round of propagation.
(ii) Existing tensor computing frameworks (e.g., Tensorflow) do not provide the tensor-matrix product operator for sparse tensors.

Inspired by the three-view drawing of engineering fields, we propose a \emph{Three-view Label Propagation} mechanism that compresses the adjacency tensor $\boldsymbol{\mathcal{A}}$ along three axes to retain maximum information from the original tensor while reducing computational complexity.
As shown in Figure \ref{fig:MRW}(a), we first separately sum the original tensor $\boldsymbol{\mathcal{A}}$ along three axes to obtain the top view $\bm A^{top}\in\mathbb{R}^{|\mathcal{R}|\times|\mathcal{E}|}$, the side view $\bm A^{side}\in\mathbb{R}^{|\mathcal{E}|\times|\mathcal{E}|}$, and the front view $\bm A^{front}\in\mathbb{R}^{|\mathcal{E}|\times|\mathcal{R}|}$.
From the perspective of inside meaning, $\bm A^{side}$, $\bm A^{front}$, and $\bm A^{top}$ represent the adjacency relations from head entity to tail entity, head entity to relation, and relation to tail entity, respectively.
Then, the labels of entities and relations are propagated according to the three views:
\begin{align}
  \bm L^{(k+1)}_e &= \bm A^{side} \bm L^{(k)}_e + \bm A^{front} \bm L^{(k)}_r\\
  \bm L^{(k+1)}_r &= \bm A^{top} \bm L^{(k)}_e
\end{align}
Intuitively, LightEA transforms the propagation process on triples into a triangular process, where each side of the triangle represents a different view (as shown in Figure \ref{fig:MRW}(b)).
In this way, labels could be effectively propagated on KGs without any trainable parameters while keeping the complexity consistent with the classical LP algorithm $O(|\mathcal{T}|d)$.
Finally, for each entity $e_i$, we concatenate the label vectors of all time-steps as the outputs:
\begin{equation}
    \bm l^{out}_{e_i} = \left[\bm l^{(0)}_{e_i}||\bm l^{(1)}_{e_i}||...||\bm l^{(k-1)}_{e_i}||\bm l^{(k)}_{e_i}\right]
\end{equation}

\subsection{Sparse Sinkhorn Iteration}
Early EA studies \cite{DBLP:conf/emnlp/WangLLZ18,DBLP:conf/semweb/SunHL17} simply calculate the Euclidean distances or cosine similarities of all entity pairs and select the closest one as the alignment result.
In this way, one entity could be aligned to multiple entities simultaneously, which violates the one-to-one constraint of EA.
To address this problem, some studies \cite{DBLP:conf/aaai/XuSFSY20, DBLP:conf/emnlp/MaoWWL21} transform the decoding process of EA into an assignment problem and achieve significant performance improvement:
\begin{equation}
  \underset{\bm P\in\mathbb{P}_{|\mathcal{E}|}}{arg\;max}{\;\left\langle \bm P, \bm S\right\rangle}_F
\end{equation}
$\bm S \in \mathbb{R}^{|\mathcal{E}|\times|\mathcal{E}|}$ is the similarity matrix of entities, and $s_{i,j} = cosine(\bm l^{out}_{e_i},\bm l^{out}_{e_j})$ in LightEA.
$\bm P$ is a permutation matrix denoting the alignment plan, where exactly one entry of $1$ in each row and column and $0$s elsewhere.
$\mathbb{P}_{|\mathcal{E}|}$ represents the set of all $|\mathcal{E}|$-dimensional permutation matrices.
$\langle\cdot\rangle_F$ represents the Frobenius inner product.

\begin{figure}
    \centering
    \includegraphics[width=0.95\linewidth]{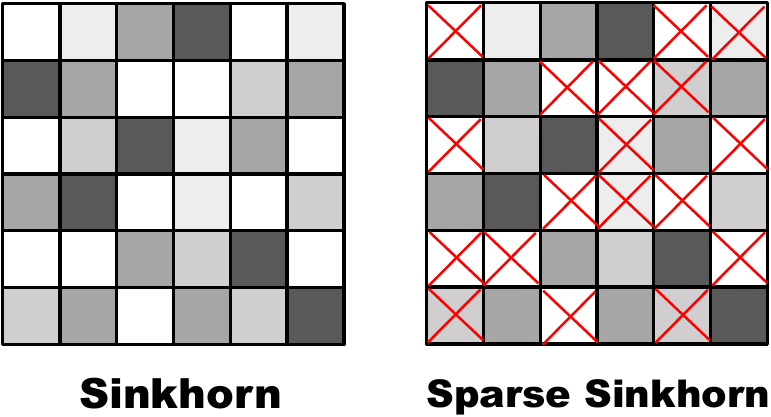}
    \caption{\emph{Sinkhorn} and \emph{Sparse Sinkhorn Iteration}.}
    \label{fig:sink}
\end{figure}
Although the \emph{Hungarian} algorithm \cite{lawler1963quadratic} could solve assignment problems accurately, its high complexity $O(|\mathcal{E}|^3)$ makes it impossible to apply in large-scale datasets.
Therefore, recent studies \cite{DBLP:conf/sigir/GeLCZG21, DBLP:conf/emnlp/MaoWWL21} propose to use the \emph{Sinkhorn iteration} \footnote{The implementation details of the \emph{Sinkhorn iteration} are listed in Appendix \ref{apd:sinkhorn}.} \cite{DBLP:conf/nips/Cuturi13} to obtain an approximate solution:
\begin{align}
\begin{split}
  &\underset{\bm P\in\mathbb{P}_{|\mathcal{E}|}}{arg\;max}{\;\left\langle \bm P,\bm S\right\rangle}_F\\
  =& \lim_{\tau\rightarrow 0^+}{\rm Sinkhorn}(\bm S/\tau)
\end{split}
\end{align}
The space-time complexity of the \emph{Sinkhorn iteration} is $O(q|\mathcal{E}|^2)$, where $q$ is the number of iteration rounds.
Even though the \emph{Sinkhorn iteration} is significantly faster than the \emph{Hungarian} algorithm, it still becomes the main bottleneck of LightEA.

In LightEA, we adopt the decoding algorithm proposed by \citet{Mao2022AnEA} and utilize \emph{Sparse Sinkhorn Iteration} to reduce the computational complexity.
Specifically, we notice that the exponential normalization of the \emph{Sinkhorn iteration} causes most smaller values in the similarity matrix $\bm S$ to be infinitely close to zero.
Even if these smaller values are removed initially, it does not significantly affect the alignment results.
Therefore, instead of calculating the similarities between all entities, LightEA only retrieves the top-$k$ nearest neighbors for each entity by \emph{Approximate Nearest Neighbor} algorithms \footnote{In LighEA, we use the FAISS framework \cite{johnson2019billion} for approximate vector retrieval.}, such as \emph{Inverted Index System} and \emph{Product Quantizer} \cite{DBLP:journals/pami/JegouDS11}.
As shown in Figure \ref{fig:sink}, the similarity matrix $\bm S$ will become sparse.
There are only $k$ non-zero elements in each row, while the others are set to zeros.
In this way, the complexity of the \emph{Sinkhorn iteration} could be reduced to $O(qk|\mathcal{E}|)$.

\begin{table*}[t]
\begin{center}
\resizebox{1\textwidth}{!}{
\renewcommand\arraystretch{0.95}
\begin{tabular}{c|ccc|ccc|ccc|ccc|ccc}
  \toprule
  \multicolumn{1}{c}{\multirow{2}{*}{Method}} & \multicolumn{3}{c}{$\rm{DBP_{ZH-EN}}$} & \multicolumn{3}{c}{$\rm{DBP_{JA-EN}}$} & \multicolumn{3}{c}{$\rm{DBP_{FR-EN}}$}& \multicolumn{3}{c}{$\rm{SRPRS_{FR-EN}}$}& \multicolumn{3}{c}{$\rm{SRPRS_{DE-EN}}$}  \\
  & H@1 & H@10 & MRR & H@1 & H@10 & MRR & H@1 & H@10 & MRR & H@1 & H@10 & MRR & H@1 & H@10 & MRR\\
  \hline
  MTransE & 0.209 & 0.512 & 0.310 & 0.250 & 0.572 & 0.360 & 0.247 & 0.577 & 0.360 &0.213& 0.447&0.290&0.107&0.248&0.160\\
  GCN-Align & 0.434 & 0.762 & 0.550 & 0.427 & 0.762 & 0.540 & 0.411 & 0.772 & 0.530 & 0.243& 0.522&0.340&0.385&0.600& 0.460\\
  RSNs&0.508&0.745&0.591&0.507&0.737&0.590&0.516&0.768&0.605&0.350&0.636& 0.440& 0.484& 0.729&0.570\\
  HyperKA&0.572&0.865&0.678& 0.564&0.865& 0.673&0.597&0.891& 0.704& -& -& -&- &-& -\\
  \textbf{LightEA-B} &\textbf{0.756}&\textbf{0.905}&\textbf{0.811}&\textbf{0.762}&\textbf{0.919}&\textbf{0.819}&\textbf{0.807}&\textbf{0.943}&\textbf{0.857}&\textbf{0.466 }&\textbf{0.746}&\textbf{0.560}&\textbf{0.594}&\textbf{0.814}&\textbf{0.670}\\
  \hline
  BootEA & 0.629 & 0.847 & 0.703 & 0.622 & 0.853 & 0.701 & 0.653 & 0.874 & 0.731 & 0.365&0.649&0.460&0.503&0.732&0.580\\
  TransEdge&0.735&0.919&0.801&0.719&0.932&0.795&0.710&0.941&0.796&0.400&0.675&0.490&0.556&0.753&0.630\\
  MRAEA &0.757&0.930&0.827&0.758&0.934&0.826&0.781&0.948&0.849&0.460&0.768&0.559&0.594&0.818&0.666\\
  Dual-AMN &0.808&\textbf{0.940}&\textbf{0.857}&0.801&\textbf{0.949}&0.855&0.840&\textbf{0.965}&0.888&0.481&\textbf{0.778}&0.568&0.614&\textbf{0.823}&\textbf{0.687}\\
  \textbf{LightEA-I} &\textbf{0.812}&0.915&0.849&\textbf{0.821}&0.933&\textbf{0.864}&\textbf{0.863}&0.959&\textbf{0.900}&\textbf{0.484}&0.769&\textbf{0.570}&\textbf{0.615}&0.817&0.685\\
  \hline
  GM-Align & 0.679 & 0.785 & - & 0.739 & 0.872 & - & 0.894 & 0.952 & - & 0.574&0.646&0.602&0.681&0.748&0.710\\
  RDGCN & 0.697 & 0.842 & 0.750 & 0.763 & 0.897 & 0.810 & 0.873 & 0.950 & 0.901  &0.672&0.767& 0.710&0.779&0.886&0.820 \\
  EASY &0.898& 0.979& 0.930 &0.943 &0.990 &0.960 &0.980 &0.998 &0.990 &0.965 &0.989 &0.970 &0.974 &0.992 &0.980\\
  SEU &0.900 &0.965 &0.924 &0.956 &0.991 &0.969 &0.988 &\textbf{0.999} &0.992 &0.982 &\textbf{0.995} &0.986 &0.983 &0.996 &0.987\\
  \textbf{LightEA-L} &\textbf{0.952}&\textbf{0.984}&\textbf{0.964}&\textbf{0.981}&\textbf{0.997}&\textbf{0.987}&\textbf{0.995}&0.998&\textbf{0.996}&\textbf{0.986}&0.994&\textbf{0.989}&\textbf{0.988 }&\textbf{0.995}&\textbf{0.991}\\
  \bottomrule
\end{tabular}
}
\caption{Performances on DBP$15$K and SRPRS. Baselines' results are from original papers or \citet{9174835}.}
\label{table:res1}
\end{center}
\end{table*}

\section{Experiments}
All experiments are conducted on a PC with an Nvidia RTX$3090$ GPU and an EPYC 7452 CPU.

\subsection{Datesets and Metrics}
To comprehensively verify the scalability, robustness, and interpretability of our proposed methods, we conduct experiments on the following four groups of datasets:

(1) \textbf{DBP15K} \cite{DBLP:conf/semweb/SunHL17} is the most commonly used EA dataset, consisting of three small-sized cross-lingual subsets.
Each subset contains $15,000$ pre-aligned entity pairs.

(2) \textbf{SRPRS} \cite{DBLP:conf/icml/GuoSH19} is a sparse dataset that includes two small-sized cross-lingual subsets.
Each subset of SRPRS also contains $15,000$ pre-aligned entity pairs but with much fewer triples.

(3) \textbf{DWY100K} \cite{DBLP:conf/ijcai/SunHZQ18} comprises two medium-sized mono-lingual subsets, each containing $100,000$ pre-aligned entity pairs and nearly one million triples.

(4) \textbf{DBP1M} \cite{DBLP:journals/pvldb/GeLCZG21} is one of the largest EA datasets so far, consisting of two cross-lingual subsets with more than one million entities and nearly ten million triples.

The detailed statistics are listed in Table \ref{table:dataset1}.
Consistent with the previous studies \cite{DBLP:conf/semweb/SunHL17,DBLP:conf/ijcai/WuLF0Y019,DBLP:conf/wsdm/MaoWXLW20}, we randomly split $30\%$ of the pre-aligned entity pairs for training and the remaining $70\%$ for testing.
Following convention \cite{DBLP:conf/ijcai/ChenTYZ17,DBLP:conf/emnlp/WangLLZ18}, we use $Hits@k$ and \emph{Mean Reciprocal Rank} (MRR) as our evaluation metrics.
The reported performances are the average of five independent runs.

\subsection{Baselines}
According to the categorization in Table \ref{tabel:rw}, we compare LightEA against the following three groups of advanced EA methods:
(1) \textbf{Basic}:
MtransE \cite{DBLP:conf/ijcai/ChenTYZ17}, GCN-Align \cite{DBLP:conf/emnlp/WangLLZ18}, RSNs \cite{DBLP:conf/icml/GuoSH19}, HyperKA \cite{DBLP:conf/emnlp/SunCHWDZ20}.
(2) \textbf{Iterative}:
BootEA \cite{DBLP:conf/ijcai/SunHZQ18}, TransEdge \cite{DBLP:journals/corr/abs-2004-13579}, MRAEA \cite{DBLP:conf/wsdm/MaoWXLW20}, Dual-AMN \cite{DBLP:conf/www/MaoWWL21}, LargeEA \cite{DBLP:journals/pvldb/GeLCZG21}, ClusterEA \cite{DBLP:journals/corr/abs-2205-10312}.
(3) \textbf{Literal}:
GM-Align \cite{DBLP:conf/acl/XuWYFSWY19}, RDGCN \cite{DBLP:conf/ijcai/WuLF0Y019}, SEU \cite{DBLP:conf/emnlp/MaoWWL21}, EASY \cite{DBLP:conf/sigir/GeLCZG21}.

To make a fair comparison against the above methods, LightEA also has three corresponding versions:
(1) LightEA-B is the basic version.
(2) LightEA-I adopts the bi-directional iterative strategy proposed by \citet{DBLP:conf/wsdm/MaoWXLW20}.
(3) LightEA-L uses the pre-trained word embeddings of translated entity names \footnote{The name translations and word embeddings are provided by \citet{DBLP:conf/acl/XuWYFSWY19}, which is consistent with follow-up studies.} as the inputs matrix and also adopts the bi-directional iterative strategy.
Same with SEU and EASY, LightEA-L is an unsupervised method that does not require any pre-aligned entity pairs.

\subsection{Hyper-parameters}
\label{sec:hyper}
Except for DBP$1$M, we use the same setting:
the dimension of hyper-sphere $d = 1,024$;
the number of \emph{Three-view Label Propagation} rounds $k = 2$.
We reserve the top-$500$ nearest neighbors for each entity in \emph{Sparse Sinkhorn Iteration}.
Following \citet{DBLP:conf/emnlp/MaoWWL21}, the number of Sinkhorn iteration rounds $q = 10$ and the temperature $\tau = 0.05$.
Due to the limitation of GPU memory, the dimension of hyper-sphere $d$ is reduced to $256$ for DBP$1$M.

\begin{table}[t]
\resizebox{\linewidth}{!}{
\renewcommand\arraystretch{0.9}
\begin{tabular}{ccccccc}
  \toprule
  \multicolumn{1}{c}{\multirow{2}{*}{Method}} & \multicolumn{3}{c}{$\rm{DWY_{DBP-WD}}$}& \multicolumn{3}{c}{$\rm{DWY_{DBP-YG}}$}  \\
  \multicolumn{1}{c}{} & H@1 & H@10 & MRR & H@1 & H@10 & MRR\\
  \hline
   MTransE &0.238&0.507&0.330&0.227&0.414&0.290\\
   GCN-Align&0.494&0.756&0.590&0.598&0.829&0.680\\
   RSNs&0.607&0.793&0.673&0.689&0.878&0.756\\
   MuGNN &0.604&0.894&0.701&0.739&0.937&0.810\\
   \textbf{LightEA-B}&\textbf{0.861}&\textbf{0.962}&\textbf{0.898}&\textbf{0.884}&\textbf{0.977}&\textbf{0.918}\\
  \hline
   BootEA &0.748&0.898&0.801&0.761&0.894&0.808\\
   TransEdge&0.788&0.938&0.824&0.792&0.936&0.832\\
   MRAEA&0.794&0.930&0.856&0.819&0.951&0.875\\
   Dual-AMN&0.869&0.969&0.908&\textbf{0.907}&\textbf{0.981}&\textbf{0.935}\\
   \textbf{LightEA-I}&\textbf{0.907}&\textbf{0.978}&\textbf{0.934}&0.902&0.980&0.929\\
  \bottomrule
\end{tabular}
}
\caption{Experimental results on DWY$100$K.}
\label{table:res2}
\end{table}

\begin{table}[t]
\resizebox{1\linewidth}{!}{
\renewcommand\arraystretch{0.9}
\begin{tabular}{ccccccc}
  \toprule
  \multicolumn{1}{c}{\multirow{2}{*}{Method}} & \multicolumn{3}{c}{$\rm{DBP1M_{FR-EN}}$}& \multicolumn{3}{c}{$\rm{DBP1M_{DE-EN}}$}  \\
  \multicolumn{1}{c}{} & H@1 & H@10 & MRR & H@1 & H@10 & MRR\\
  \hline
    LargeEA-G &0.051 &0.134& 0.080 & 0.034& 0.095& 0.050 \\
    LargeEA-R &0.094& 0.215& 0.130&  0.064 &0.150 &0.090 \\
    LargeEA-D &0.105 &0.219 &0.150&  0.066&0.147& 0.090 \\
    ClusterEA-G &0.100 &0.245 &0.150 &0.069 &0.177 &0.110 \\
    ClusterEA-R &0.260& 0.456& 0.320 & 0.250 &0.450& 0.320 \\
    ClusterEA-D &0.281& \textbf{0.474}& \textbf{0.350}& 0.288& \textbf{0.488} &\textbf{0.350} \\
   \hline
   \textbf{LightEA-B}&0.262&0.450& 0.318&0.258&0.457&0.316\\
   \textbf{LightEA-I}&\textbf{0.285}&0.468&0.345&\textbf{0.289}&0.479&0.347\\
  \bottomrule
\end{tabular}
}
\caption{Experimental results on DBP$1$M\protect\footnotemark.}
\label{table:res3}
\end{table}
\footnotetext{The results of baselines are from ClusterEA\cite{DBLP:journals/corr/abs-2205-10312}. To solve the name bias issue, ClusterEA removes parts of entities and all literal information while retaining all the triples. Therefore, LightEA-L cannot run on this dataset.}

\subsection{Main Experiments}
Table \ref{table:res1} lists the experimental results of LightEA on DBP$15$K and SRPRS.
Table \ref{table:res2} and Table \ref{table:res3} list the results on DWY$100$K and DBP$1$M, respectively.

\noindent
\textbf{DBP15K}.
Compared to the basic EA baselines, LightEA-B has significant improvements on all metrics.
The main reason is that these basic EA methods were proposed earlier, while most advanced methods adopts iterative strategies for better performance.
For LightEA-I, the bi-directional iterative strategy improves the performance by more than $5\%$ on $Hits@1$ and $4\%$ on MRR.
Compared to Dual-AMN, the state-of-the-art structure-based EA method, LightEA-I achieves comparable results that are slightly better on $Hits@1$ and weaker on $Hits@10$.
Among literal-based EA methods, LightEA-L consistently achieves the best performances.
Especially for $\rm DBP_{ZH-EN}$, LightEA-L beats SEU by more than $5\%$ on $Hits@1$.

\noindent
\textbf{SRPRS}.
The performance rankings on SRPRS are pretty similar to those on DBP$15$K.
LightEA-B and LightEA-L defeat all the competitors, while LightEA-I achieves similar performances to Dual-AMN.
The main difference is that the impact of iterative strategies is significantly weakened on SPRPS, in which the performance gain is less than $2\%$ on $Hits@1$ and $1\%$ on MRR.
That is mainly because the structural information of SRPRS is particularly sparse, causing iterative strategies to fail to generate high-quality semi-supervised data.

\noindent
\textbf{DWY100K}.
For DWY$100$K, LightEA maintains competitive performance.
On $\rm DWY_{DBP-WD}$, LightEA outperforms Dual-AMN by $4\%$ on $Hits@1$ and $3\%$ on MRR while having a comparable result on $\rm DWY_{DBP-YG}$.
According to \citet{9174835}, the names of equivalent entities in DWY$100$K are almost identical, and using the edit distance could easily achieve the ground-truth performance.
Therefore, Table \ref{table:res2} does not list the results of literal methods.

\noindent
\textbf{DBP1M}.
As one of the largest EA datasets so far, DBP1M poses a severe challenge to scalability, and most EA methods cannot directly run on this dataset.
As shown in Table \ref{table:res3}, the experimental results show that LightEA still achieves similar performances to the state-of-the-art method on this challenging dataset.
LargeEA and ClusterEA are two acceleration frameworks for GNN-based EA methods, consisting of mini-batch graph samplers, efficient losses, etc.
"-G," "-R," and "-D" represent using GCN-Align, RREA, and Dual-AMN as the backbone networks, respectively.

\begin{table}
\begin{center}
\resizebox{1\linewidth}{!}{
\renewcommand\arraystretch{0.9}
\begin{tabular}{ccccc}
  \toprule
  \textbf{Method}&\textbf{DBP15K}&\textbf{SRPRS}&\textbf{DWY100K}&\textbf{DBP1M}\\
  \hline
  Dual-AMN&69&57&2,237&-\\
  LargeEA-D&26&23&227&2,119\\
  ClusterEA-D&43&36&389&1,503\\
  \hline
  LightEA-B&2.8&2.2&15.4&97\\
  LightEA-I&7.1&6.5&34.5&228\\
  LightEA-L&14.8&11.2&-&-\\
  \bottomrule
\end{tabular}
}
\end{center}
\caption{Time costs on all datasets (seconds). }\label{tabel:time}
\end{table}

\noindent
\textbf{Time cost}.
Table \ref{tabel:time} reports the time costs of three variants of LightEA.
Due to the space limitation, we only list out the time costs of Dual-AMN with two acceleration frameworks as the baselines.
For the results of other EA methods, please refer to Table \ref{tabel:timeapp} in Appendix C.
LargeEA and ClusterEA effectively accelerate Dual-AMN, achieving ten times speedups on DBP$1$M.
However, GNN-based methods are still stuck with multiple rounds of forward and backward propagation, while LightEA only requires one round of label propagation and no need for training.
With similar performances, the time consumption of LightEA-B is less than one-seventh of Dual-AMN variants.
Besides, introducing iterative strategies and literal features significantly increases time consumption while also improving performance.

\begin{figure*}
    \centering
    \includegraphics[width=0.95\textwidth]{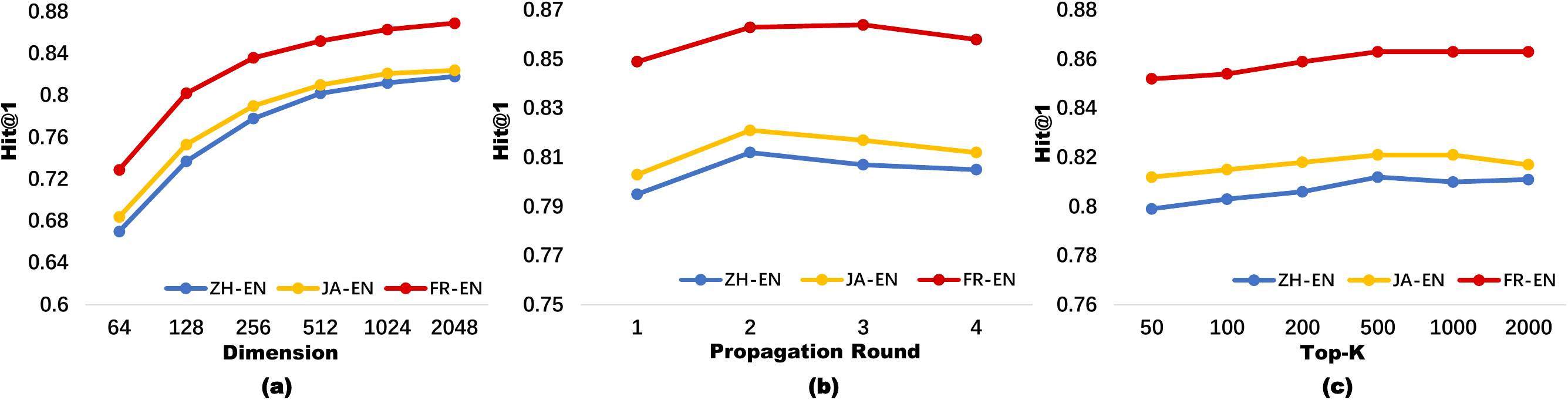}
    \caption{Hyper-parameter experiments of LightEA-I on DBP$15$K.}
    \label{fig:hyper}
\end{figure*}

\subsection{Hyper-parameters}
We design extensive hyper-parameter experiments to investigate the behaviors of LightEA in various situations.
Due to the space limitation, Figure \ref{fig:hyper} only shows the results of LightEA-I on DBP$15$K.
Appendix \ref{sec:hyperapp} shows the results on other datasets.

\noindent
\textbf{Dimension}.
To alleviate the problem that one-hot label vectors being over-sparse, LightEA proposes to replace them with independent random vectors on high-dimensional hyper-spheres.
Figure \ref{fig:hyper}(a) shows the $Hits@1$ curves with different dimensions $d$.
Clearly, there are significant diminishing marginal effects on increasing dimensions.
When the dimension $d>1,024$, the performance gain becomes limited.
Actually, the reason behind diminishing marginal effects has been told by Lemma \ref{lemma:1} --- the probability of two random drawn vectors "conflicting" with each other (i.e., $\langle\bm x,\bm y\rangle > \epsilon$) decreases exponentially as the dimension $d$ increases.

\noindent
\textbf{Propagation Round}.
Figure \ref{fig:hyper}(b) shows the performances with different numbers of propagation rounds.
Similar to the network depth of GNNs, more propagation rounds also lead to the over-smoothing problem.
When the number of rounds $q=2$, LightEA-I achieves the best performance and then begins to decline.
This phenomenon indicates a high correlation between label propagation and GNNs from the side.

\noindent
\textbf{Top-K}.
In \emph{Sparse Sinkhorn Iteration}, LightEA only retrieves the top-$k$ nearest neighbors for each entity instead of calculating the distances between all pairs.
As expected, Figure \ref{fig:hyper}(c) shows that removing smaller values from the similarity matrix hardly affects the alignment performances.
The trick of \emph{Sparse Sinkhorn Iteration} could reduce the space-time complexity with limited losses.

Besides hyper-parameters, we further design more auxiliary experiments, such as ablation experiments.
Due to the space limitation, these experiments are listed in Appendix \ref{sec:pre}, \ref{sec:able}, and \ref{sec:openEA}.

\subsection{How to Interpret the Results}
\label{sec:interpret}
Existing GNN-based EA methods inherit poor interpretability from neural networks.
It is hard to explain the inside meaning of each dimension of entity features and why two entities are aligned.
Therefore, the output features are only used to calculate the similarities.
Different from GNN-based methods, we could clearly interpret the alignment results of LightEA by the following steps:

(1) Remove \emph{Random Orthogonal Label}.
After removing this component, the meaning of each dimension of label vectors becomes clear and realistic.
The $x$-$th$ dimension of $\bm l_{e_i}^{(k)}$ represents the relevance score between $e_i$ and the $x$-$th$ pre-aligned entity pair after $k$ rounds of propagation.
If two entities are equivalent, their distributions of relevance scores should also be similar.

(2) Trace the propagation:
Since we have known the meaning of each dimension, we could trace the propagation process at each time-step to investigate why two entities are aligned.

However, removing the \emph{Random Orthogonal Label Generation} will make the label matrices extremely large and sparse, and the \emph{Three-view Label Propagation} can only run on the small-sized sub-graphs.
Therefore, this strategy only fits for interpreting a limited number of entities.

\begin{figure*}[t]
    \centering
    \includegraphics[width=\textwidth]{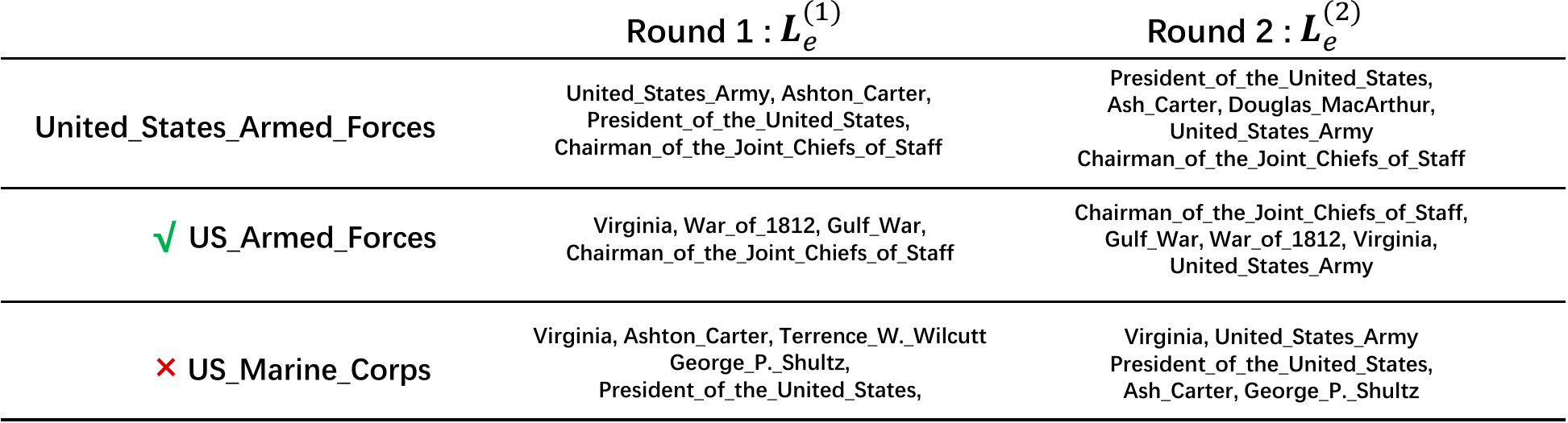}
    \caption{An example of tracing the propagation process and interpreting the alignment results.}
    \label{fig:wrong}
\end{figure*}

\subsection{An Example of Tracing the Wrong Case}
\label{sec:wrong_case}
Figure \ref{fig:wrong} shows an example of tracing the propagation process and interpreting the alignment results.
In this case, \emph{United\_States\_Armed\_Forces} is incorrectly aligned with \emph{US\_Marine\_Corps}, and the correct answer is \emph{US\_Armed\_Forces}.

As described in last section, we first remove the \emph{Random Orthogonal Label}, run the \emph{Three-view Label Propagation}, and obtain the sparse label matrices $\bm L^{(1)}_e$ and $\bm L^{(2)}_e$.
Then, we separately sort the label vectors $\bm l^{(1)}_{e_i}$ and $\bm l^{(2)}_{e_i}$ of each entity to get the dimension indexes of top-$5$ non-zero elements.
Finally, we look up the corresponding names of dimension indexes in the entity name list and show them in Figure \ref{fig:wrong}.
Apparently, the noise of graph structure causes this wrong case.
Compared to the correct entity, the incorrect entity has a more similar neighborhood structure to \emph{United\_States\_Armed\_Forces}.
There are five common elements between the wrong alignment pair, while only three between the right alignment pair.

\section{Conclusion}
In this paper, we reinvent the \emph{Label Propagation} algorithm to effectively run on KGs and propose a non-neural EA framework --- LightEA.
According to the experiments on public datasets, LightEA has impressive scalability, robustness, and interpretability.
With a mere tenth of time consumption, LightEA achieves comparable results to state-of-the-art methods across all datasets.

\section*{Limitations}
Although we have demonstrated that LightEA has impressive scalability, robustness, and interpretability on multiple public datasets with different scales, there are still three limitations that should be addressed in the future:

(1) In LightEA, good interpretability and high efficiency cannot coexist.
If not removing the \emph{Random Orthogonal Label}, LightEA's interpretability will be significantly weakened.
How to balance the interpretability and efficiency is our future work.

(2) Theoretically, LightEA has high parallel efficiency that could obtain linear speedup with multiple GPUs.
However, we do not have enough devices to verify this advantage, and all the experiments in this paper run with a single RTX3090.
We will purchase more devices to complete these missing experiments in the future.

(3) Currently, we implement LightEA via Tensorflow.
Since LightEA is a non-neural algorithm without any trainable parameters, a complex deep learning framework would be redundant and inefficient.
In the future, we will refactor LightEA with CUDA and C++ to further improve efficiency.

\section*{Ethics Statement}
To the best of our knowledge, this work does not involve any discrimination, social bias, or private data.
All the datasets are open-source and constructed from open-source KGs such as Wikidata, YAGO, and DBpedia.
Therefore, we believe that our study complies with the \href{https://www.aclweb.org/portal/content/acl-code-ethics}{ACL Ethics Policy}.
In addition, introducing literal features may cause concerns about the misuse of user-generated content.
LightEA can avoid this concern by using pure structural information for entity alignment.

\section*{Ethics Statement}
We appreciate the support from National Natural Science Foundation of China with the Research Project on Human Behavior in Human-Machine Collaboration (Project Number: 72192824) and the support from Pudong New Area Science \& Technology Development Fund (Project Number: PKX2021-R05).

\bibliography{custom}
\bibliographystyle{acl_natbib}
\clearpage
\appendix

\section{Sinkhorn Iteration}
\label{apd:sinkhorn}
\begin{align}
\begin{split}
    {\rm Sink}^{(0)}(\bm S) &= \exp(\bm S),\\
    {\rm Sink}^{(q)}(\bm S) &= {\mathcal N}_c({\mathcal N}_r({\rm Sink}^{(q-1)}(\bm S))),\\
    {\rm Sinkhorn}(\bm S) &= \lim_{q\rightarrow\infty}{\rm Sink}^{(q)}(\bm S).
\end{split}
\end{align}
here {\small${\mathcal N}_r(\bm S)$$=$$ \bm S \varoslash (\bm S \bm 1_{N}\bm 1_{N}^T)$} and {\small ${\mathcal N}_c $$=$$ \bm S \varoslash (\bm 1_{N}\bm 1_{N}^T\bm S)$} are the row and column-wise normalization operators of a matrix, $\varoslash$ represents the element-wise division, and $\bm 1_{N}$ is a column vector of ones.

\section{Datasets}
\begin{table}[h]
\begin{center}
\resizebox{\linewidth}{!}{
\renewcommand\arraystretch{0.9}
\begin{tabular}{p{2cm}c|cccccc}
\toprule
\multicolumn{2}{c|}{Datasets} & $|\mathcal{E}|$ & $|\mathcal{R}|$  & $|\mathcal{T}|$\\
\toprule
\multirow{2}{1.3cm}{$\rm{DBP_{ZH-EN}}$} & ZH & 19,388 & 1,701& 70,414\\
& EN & 19,572 & 1,323 & 95,142 \\
\multirow{2}{1.3cm}{$\rm{DBP_{JA-EN}}$} & JA & 19,814 & 1,299 & 77,214\\
& EN & 19,780 & 1,153  & 93,484 \\
\multirow{2}{1.3cm}{$\rm{DBP_{FR-EN}}$} & FR & 19,661 & 903 & 105,998\\
& EN & 19,993 & 1,208 & 115,722  \\
\hline
\multirow{2}{1.3cm}{$\rm{DWY_{DBP-YG}}$} & DBP & 100,000 &  302  & 428,952\\
& YG & 100,000 & 31  &  502,563 \\
\multirow{2}{1.3cm}{$\rm{DWY_{DBP-WD}}$} & DBP & 100,000 &  330 & 463,294\\
& Wiki & 100,000 &  220  &  448,774  \\
\hline
\multirow{2}{1.3cm}{$\rm{SRPRS_{FR-EN}}$} & FR & 15,000 & 177& 33,532\\
& EN & 15,000 & 221& 36,508 \\
\multirow{2}{1.3cm}{$\rm{SRPRS_{DE-EN}}$} & DE & 15,000 & 120 & 37,377\\
& EN & 15,000 & 222 & 38,363  \\
\hline
\multirow{2}{1.3cm}{$\rm{DBP1M_{FR-EN}}$} & FR & 1,365,118 & 380& 2,997,457\\
& EN & 1,877,793 & 603& 7,031,172 \\
\multirow{2}{1.3cm}{$\rm{DBP1M_{DE-EN}}$} & DE & 1,112,970 & 241 & 1,994,876\\
& EN & 1,625,999 & 597 & 6,213,639  \\
\bottomrule
\end{tabular}
}
\end{center}
\caption{Statistical data of all datasets.}\label{table:dataset1}
\end{table}

\section{Time Costs}
\begin{table}[h]
\begin{center}
\resizebox{0.9\linewidth}{!}{
\renewcommand\arraystretch{0.85}
\begin{tabular}{cccc}
  \toprule
  \textbf{Method}&\textbf{DBP15K}&\textbf{SRPRS}&\textbf{DWY100K}\\
  \toprule
  MTransE  &6,467&3,355&70,085\\
  GCN-Align &103&87&3,212\\
  RSNs &7,539&2,602&28,516\\
  MuGNN&3,156&2,215&47,735\\
  KECG &3,724&1,800&125,386\\
  \midrule
  BootEA &4,661&2,659&64,471\\
  NAEA &19,115&11,746&-\\
  TransEdge&3,629&1,210&20,839\\
  MRAEA &3,894&1,248&23,275\\
  \midrule
  GM-Align &26,328&13,032&459,715\\
  RDGCN &6,711&886&-\\
  HMAN &5,455&4,424&31,895\\
  HGCN &11,275&2,504&60,005\\
  \bottomrule
\end{tabular}
}
\end{center}
\caption{Time costs of existing EA methods (seconds). Because we do not have enough time and devices to run all these methods by ourselves, these time costs are from the summary \cite{9174835} for reference only.}\label{tabel:timeapp}
\end{table}

\section{Hyper-parameter}
\label{sec:hyperapp}
\begin{figure*}[t]
    \centering
    \includegraphics[width=\textwidth]{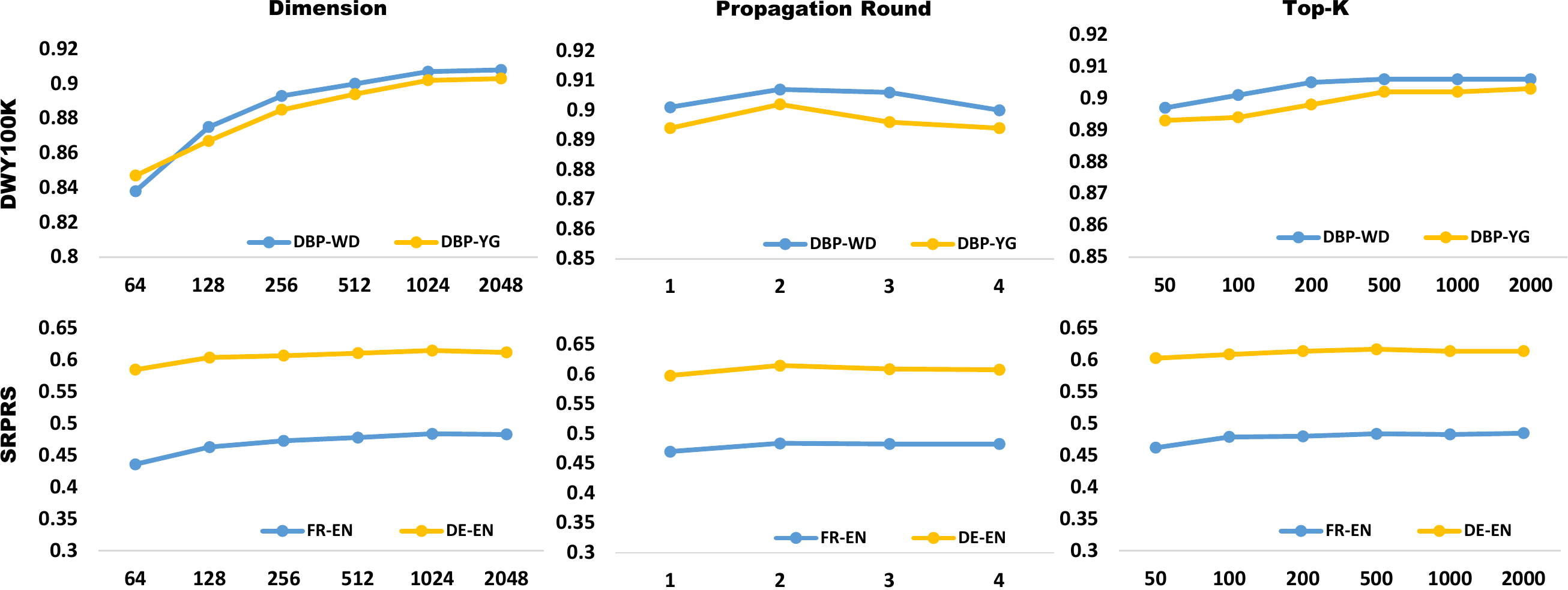}
    \caption{Hyper-parameter experiments of LightEA-I on DWY$100$K and SRPRS.}
    \label{fig:hyperapp}
\end{figure*}

Figure \ref{fig:hyperapp} shows the remaining hyper-parameter experiments on DWY$100$K and SRPRS.
The performance curves on DWY$100$K are quite similar to those on DBP$15$K because the graph densities of these two datasets are close.
On SRPRS, we notice that these three hyper-parameters do not significantly impact performance because SRPRS is a highly sparse and small-sized dataset.
Due to the limitation of GPU memory, we do not experiment on DBP$1$M, and we will try our best to supplement these experiments in the future.

\section{Pre-aligned Ratio}
\label{sec:pre}
\begin{figure*}[t]
    \centering
    \includegraphics[width=\textwidth]{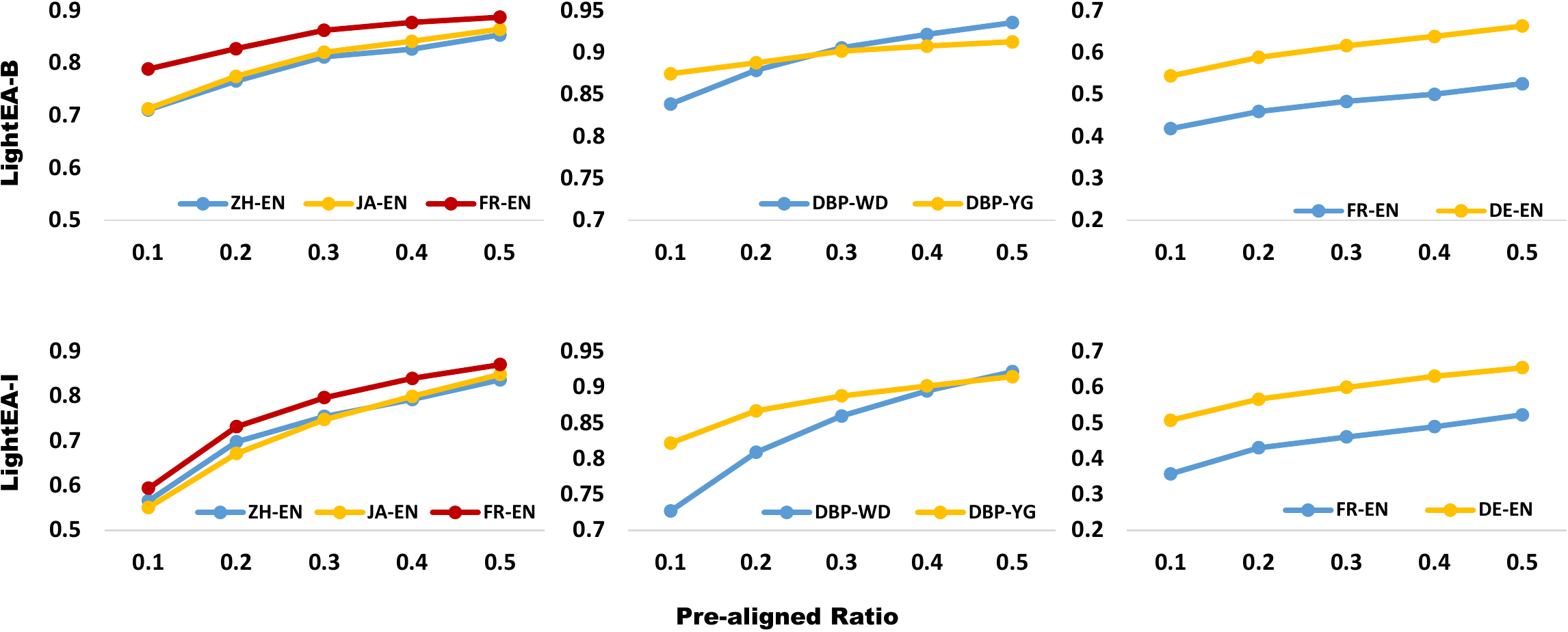}
    \caption{\emph{Hits@1} performances with different pre-aligned ratios on DBP$15$K, DWY$100$K, and SRPRS.}
    \label{fig:pre-aligned}
\end{figure*}

In practice, manually annotating pre-aligned entity pairs is labor-consuming, especially for large-scale KGs.
Therefore, we hope the proposed EA methods could maintain decent performances with limited pre-aligned entity pairs.
To investigate the behaviors of LightEA with different pre-aligned ratios, we set the ratios from $10\%$ to $50\%$.
Figure \ref{fig:pre-aligned} shows the $Hits@1$ performances of LightEA-B and LightEA-I on DBP$15K$, DWY$100$K, and SRPRS.
Even if only 10\% of the entity pairs are reserved as the training data, LightEA still achieves decent performances.
Besides, the iterative strategy effectively improves performance and alleviates the need for pre-aligned pairs on all datasets.

\section{Ablation Study}
\label{sec:able}
We design the following ablation experiment to demonstrate the effectiveness of each component of LightEA:
(i) Remove the front view $\bm A^{front}$ and the top view $\bm A^{top}$. Thus, the \emph{Three-view Label Propagation} will be degraded into the classical LP.
(ii) Remove the \emph{Sparse Sinkhorn Iteration} and select the closest entity as the alignment result.
Due to the limitation of GPU memory, the \emph{Random Orthogonal Label} cannot be removed.
Table \ref{table:ablation} shows the ablation results.
Obviously, both two components are necessary, and removing either one will cause significant performance degradation.

\begin{table}[b]
\renewcommand\arraystretch{1.2}
\resizebox{1\linewidth}{!}{
\begin{tabular}{lcccccc}
\toprule
\multirow{2}{* }{Method} & \multicolumn{2}{c}{$\rm{DBP_{ZH-EN}}$} & \multicolumn{2}{c}{$\rm{DBP_{JA-EN}}$} & \multicolumn{2}{c}{$\rm{DBP_{FR-EN}}$}\\
& Hits@1 & MRR & Hits@1 & MRR & Hits@1 & MRR\\
\toprule
 LightEA-I &0.812& 0.849&0.821 &0.864&0.863& 0.900\\
 -\emph{Three-view}&0.537& 0.588&0.606 & 0.580&0.574& 0.623\\
 -\emph{Sinkhorn}&0.664& 0.739&0.638 &0.722&0.666& 0.748\\
\bottomrule
\end{tabular}
}
\caption{Ablation study of LightEA.}
\label{table:ablation}
\end{table}

\begin{table*}[t]
    \centering
    \includegraphics[width=\textwidth]{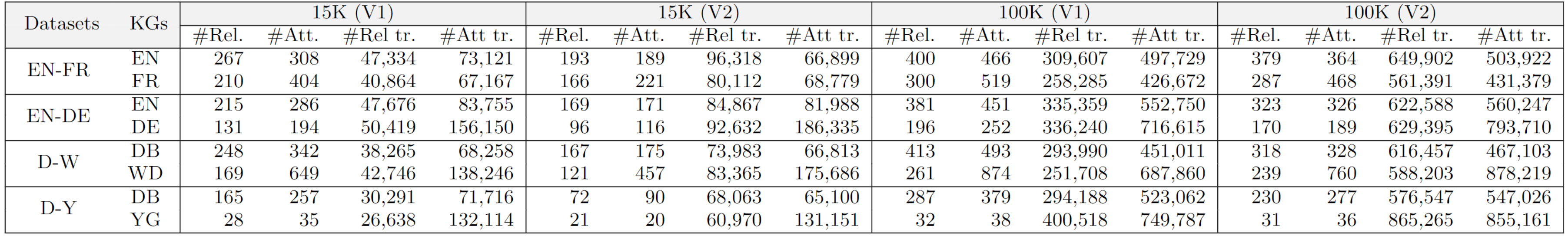}
    \caption{Statistical data of the OpenEA benchmark.}
    \label{table:openea}
\end{table*}

\clearpage
\clearpage

\section{OpenEA Benchmark (v2.0)}
\label{sec:openEA}

\begin{table}[h]
\resizebox{1\linewidth}{!}{
\renewcommand\arraystretch{1.1}
\begin{tabular}{ccccccc}
  \toprule
  \multicolumn{1}{c}{\multirow{2}{*}{Dataset}} & \multicolumn{3}{c}{\textbf{LightEA-B}}& \multicolumn{3}{c}{\textbf{LightEA-I}}  \\
  \multicolumn{1}{c}{} & H@1 & H@10 & MRR & H@1 & H@10 & MRR\\
  \hline
    EN-FR-15K-V1 & 0.607 & 0.867 & 0.697 & 0.670 & 0.895 & 0.748\\
    EN-FR-15K-V2 & 0.826 & 0.960 & 0.875 & 0.913 & 0.986 & 0.941\\
    EN-FR-100K-V1 & 0.462 & 0.714 & 0.544 & 0.507 & 0.736 & 0.581\\
    EN-FR-100K-V2 & 0.765 & 0.917 & 0.819 & 0.830 & 0.943 & 0.871\\
    \hline
    EN-DE-15K-V1 & 0.760 & 0.938 & 0.821 & 0.781 & 0.947 & 0.840\\
    EN-DE-15K-V2 & 0.919 & 0.974 & 0.939 & 0.951 & 0.987 & 0.965\\
    EN-DE-100K-V1 & 0.581 & 0.793 & 0.650& 0.604 & 0.805 & 0.670\\
    EN-DE-100K-V2 & 0.822 & 0.916 & 0.855 & 0.863 & 0.938 & 0.890\\
    \hline
    D-W-15K-V1 & 0.663 & 0.867 & 0.737& 0.732 & 0.902 & 0.796\\
    D-W-15K-V2 & 0.924 & 0.990 & 0.949 & 0.951 & 0.995 & 0.968\\
    D-W-100K-V1 & 0.588 & 0.796 & 0.659 & 0.642 & 0.833 & 0.707\\
    D-W-100K-V2 & 0.874 & 0.962 & 0.906 & 0.926 & 0.983 & 0.947\\
    \hline
    D-Y-15K-V1 & 0.770 & 0.891 & 0.817 & 0.826 & 0.927 & 0.864\\
    D-Y-15K-V2 & 0.976 & 0.995 & 0.983 & 0.976 & 0.996 & 0.983\\
    D-Y-100K-V1 & 0.758 & 0.916 & 0.811 & 0.781 & 0.931 & 0.832\\
    D-Y-100K-V2 & 0.961 & 0.990 & 0.972 & 0.977 & 0.996 & 0.984\\
  \bottomrule
\end{tabular}
}
\caption{Experimental results on OpenEA (v2.0).}
\label{table:res4}
\end{table}

To make a fair and realistic comparison, OpenEA \cite{DBLP:journals/pvldb/SunZHWCAL20} constructs a well-designed evaluation benchmark.
As shown in Table \ref{table:openea}, this benchmark contains two cross-lingual settings extracted from multi-lingual DBpedia (English-to-French and English-to-German) and two mono-lingual settings among popular KGs (DBpedia-to-Wikidata and DBpedia-to-YAGO).
Each setting has two scales with $15$K and $100$K entity pairs, respectively.
Besides, each subset is further divided into two versions with different densities.
V$1$ represents the sparse version, and V$2$ represents the dense version.
The average degree of the V$1$ datasets is about half of the V$2$ datasets.

Recently, \citet{DBLP:journals/pvldb/SunZHWCAL20} \url{} released a new version (v2.0) of the OpenEA benchmark, where the URIs of DBpedia and YAGO entities are encoded to resolve the name bias issue.
They strongly recommend using the v2.0 benchmark for evaluating EA methods, such that the results can better reflect the robustness of these methods in real-world situations.
Since this benchmark was released recently, the performances of most baselines on this benchmark are still unclear, and we do not have enough time and devices to reproduce all the baselines on this benchmark.
Therefore, Table \ref{table:res4} only reports the performances of our proposed method to offer a reference for follow-up research.
Here, we follow the data splits in OpenEA, where 20\% of alignment entity pairs are for training, 10\% for validation, and 70\% for testing.
The setting of hyper-parameters follows Section \ref{sec:hyper}.

\end{document}